\documentclass[10pt,twocolumn]{article} 
\usepackage{latex8}
\usepackage{times}
\usepackage{graphicx}

\title{CRL at NTCIR2}

\author{Masaki Murata, Masao Utiyama, Qing Ma, Hiromi Ozaku, Hitoshi Isahara\\
Communications Research Laboratory\\
  2-2-2 Hikaridai Seika-cho, Soraku-gun,\\
  Kyoto, 619-0289 Japan\\
  {\tt\{murata,mutiyama,qma,romi,isahara\}@crl.go.jp}
}

\newcommand{\score}{\mathit{score}}
\newcommand{\Score}{\mathit{Score}}
\newcommand{\tf}{\mathit{tf}}
\newcommand{\idf}{\mathit{idf}}
\newcommand{\D}{\mathcal{D}}
\newcommand{\Var}{\mbox{Var}}

\newcommand{\rel}{\mathit{rel}}

\begin{document}

\maketitle
\thispagestyle{empty}

\begin{abstract}
  We have developed systems of two types for NTCIR2. 
  One is an enhenced version of the system we developed for
  NTCIR1 and IREX. It submitted retrieval results for JJ and CC
  tasks.  A variety of parameters were tried with the system. 
  It used such characteristics of newspapers as locational
  information in the CC tasks. The system got good results for
  both of the tasks. The other system is a
  portable system which avoids free parameters as much as
  possible. The system submitted retrieval results for JJ, JE,
  EE, EJ, and CC tasks. The system automatically determined the
  number of top documents and the weight of the original query
  used in automatic-feedback retrieval. It also determined
  relevant terms quite robustly. For EJ and JE tasks, it used
  document expansion to augment the initial queries. It
  achieved good results, except on the CC tasks.  

\end{abstract}

\section{Introduction}
\label{sec:intro}

We have developed two systems for the second NTCIR Workshop's
information retrieval (IR) tasks.  

One is an enhanced
version of the system that was used for the first NTCIR
Workshop's IR tasks
\cite{murata99:_infor_retriev_based_stoch_model} and the IREX
Workshop's IR tasks
\cite{murata99:_infor_retriev_based_stoch_model_irex}. We call
this System A.
The other is a newly developed
system in which free parameters are avoided as much as
possible. We call this System B.\footnote{System A was developed mainly 
by the first author, and System B was developed mainly 
by the second author.}

System A participated
in tasks set in Japanese and Chinese (JJ and CC). It achieved high
average precisions on both tasks. 
System B participated in tasks set in Japanese, English, and
Chinese (JJ, JE, EE, EJ, and CC). It achieved high average
precisions on the JJ, EE, JE, and EJ tasks.  

Although the two systems participated in some of the same tasks,
the details of the system implementations are rather different. Thus, we
describe the two systems separately, focusing on particular
tasks; i.e., we describe System-A in the context of CC tasks and 
describe System-B in the context of JJ, EE, JE,
and EJ tasks. 

\section{Chinese IR Tasks}
\label{sec:c}
In this section, we describe System A in the context of CC tasks. 
System A participated in JJ tasks\footnote{System A 
particpated in the long-query and short-query JJ tasks. 
The best average precisions of the two tasks 
in terms of A judgement were 0.4082 (CRL20) and 0.3730 (CRL16), 
and the best average R-precisions 
were 0.4210 (CRL20) and 0.3866 (CRL27). 
These results are also good. Strings in parentheses indicate 
system ids in the NTCIR contest. 
Examination of System A's performance on JJ tasks is 
the subject of a forthcoming publication.} and CC tasks, 
and achieved particularly good results on the CC tasks. 
This reason is that the types of documents used in the CC tasks 
were very different from those used in the JJ tasks. 
While the JJ tasks involved retrieval from a 
database of academic conference papers, 
the CC tasks involved retrieval from a database 
of newspaper articles. 
System A\footnote{System A is based on the system 
we entered in the IREX contest. 
In the IREX contest, articles 
in a database of newspapers database were used 
as the test collection. System A achieved good results in 
the IREX contest, too \cite{murata99:_infor_retriev_based_stoch_model_irex,iral2000}.} takes advantage of such characteristics of newspapers as 
the title or the first sentence of the body of an article 
in a newspaper often indicating the article's subject. 
We thus expected System A to be effective on the CC tasks. 
In the following sections, we give a detailed description 
of System A and report on the experimental results of System A's 
application to the CC tasks. 

\subsection{Outline of System A}
\label{sec:outline_of_systemB}

System A uses 
Robertson's 2-poisson model \cite{robertson92:_some_simpl_effec_approx_poiss} which is one kind of 
probabilistic approach. 
In Robertson's method, 
each document's score is 
calculated 
by using the following equation.\footnote{This equation 
is BM11, which corresponds to BM25 in the case of $b = 1$ \cite{robertson_trec3}.} 
The documents that obtain high scores are then output as retrieval results. 
($Score(d,q)$ below is the score of a document $d$ 
against a query $q$.) 

{\footnotesize
\begin{eqnarray}
  \label{eqn:robertson}
&  Score(d,q) = \displaystyle \sum_{\begin{minipage}[h]{0.8cm}
    term $t$ 
    
    in $q$
    \end{minipage}} & \hspace*{-0.3cm} \left(  \displaystyle \frac{tf(d,t)}{\displaystyle tf(d,t) + k_{t} \frac{length(d)}{\Delta}} \ \times \ log\frac{N}{df(t)} \right. \nonumber \\
& &  \left. \times \displaystyle \frac{tf_{q}(q,t)}{tf_{q}(q,t) + kq} \right)
\end{eqnarray}}
where $t$ indicates a term that appears in a query. 
$tf(d,t)$ is the frequency of $t$ in a document $d$, 
$tf_{q}(q,t)$ is the frequency of $t$ in a query $q$, 
$df(t)$ is the number of the documents in which $t$ appears, 
$N$ is the total number of documents, 
$length(d)$ is the length of a document $d$, and 
$\Delta$  is the average length of the documents. 
$k_{t}$ and $k_{q}$ are constants which are set 
according to the results of experiments. 

In this equation, 
we call {\footnotesize $ \displaystyle \frac{tf(d,t)}{\displaystyle tf(d,t) + k_{t} \frac{length(d)}{\Delta}}$}
the TF term, (abbr. $TF(d,t)$), 
{\footnotesize $log\frac{N}{df(t)}$} the IDF term, (abbr. $IDF(t)$), 
and {\footnotesize $\frac{tf_{q}(q,t)}{tf_{q}(q,t) + kq}$}
the TF$_{q}$ term (abbr. $TF_{q}(q,t)$). 

In System A, several terms are added to extend 
this equation, and its method is expressed by the following equation. 

{\footnotesize
\begin{eqnarray}
&& Score(d.q)  = \displaystyle K_{\tiny category}(d) \left\{ \displaystyle \sum_{\begin{minipage}[h]{0.8cm}
      term $t$

      in $q$
\end{minipage}} \left( TF(d,t) \times IDF(t)  \right.\right. \nonumber \\
&&\hspace*{0.7cm} \left. \times TF_{q}(q,t) \times K_{\tiny location}(d,t) \times \left(log\frac{Nq}{qf(t)}\right)^{k_{Nq}}\right) \nonumber \\
&&\hspace*{0.7cm} \left. + \displaystyle \frac{length(d)}{length(d) + \Delta} \right\}
  \label{eqn:score}
\end{eqnarray}}

The TF, IDF and TF$_{q}$ terms in this equation are 
identical to those in Eq. (\ref{eqn:robertson}). 
The value of the term {\footnotesize $\frac{length}{length + \Delta}$} 
increases with the length of the document. 
This term is introduced because 
if all of the other information is exactly the same, 
the longer document is more likely to include 
content that is a relevant response to the query. 
$Nq$ is the total number of queries and $qf(t)$ is 
the number of queries in which $t$ occurs. 
Those terms which occur more frequently in queries 
are more likely to be stop words such as ``documents'' and ``thing.'' 
We decrease the scores of stop words by using $log\frac{Nq}{qf(t)}$. 
$K_{\tiny category}$ and $K_{\tiny location}$ 
are extended numerical terms that are introduced to improve precision 
of results. 
$K_{\tiny category}$ uses the category information of the document 
found in newspapers, such as 
the economic or political pages. 
$K_{\tiny location}$ uses the location of the term within the document. 
If the term is in the title or 
at the beginning of the body of the document, 
it is given a higher weighting. 
In the next section,
we explain these extended numerical terms in detail. 

\subsection{Extended numerical terms}

We use the two extended numerical terms 
$K_{\tiny location}$ and $K_{\tiny category}$ 
as shown in Eq. (\ref{eqn:score}). 
In this section, they are explained in detail. 

\begin{enumerate}
\item 
\underline{Location information ($K_{\tiny location}$)} 

In general, 
the title or the first sentence of the body of a document 
in a newspaper indicates its subject. 
Therefore, the precision of information retrieval 
can be improved by assigning more weight to the terms 
from these two locations. 
This is achieved by $K_{\tiny location}$ which adjusts the weight 
on a term the basis of whether or not it appears 
at the beginning of the document. 
If a term is in the title or 
at the beginning of the body, 
it is given a high weighting. 
Otherwise, it is given a low weighting. 
$K_{\tiny location}$ is expressed as follows: 

{\footnotesize
\begin{eqnarray}
\hspace*{-0.5cm}
K_{\tiny location}(d,t) = \left\{ 
  \begin{array}[h]{l}
k_{\tiny location,1}  \\
\mbox{(when a term $t$ occurs in the title of}\\
\mbox{a document $d$),}\\[0.2cm]
1 + k_{\tiny location,2} \displaystyle \frac{(length(d) - 2*P(d,t))}{length(d)}\\
\mbox{(otherwise)}
  \end{array}\right.\hspace*{-0.5cm}
\label{eqn:ichi}
\end{eqnarray}}

$P(d,t)$ is the location of a term $t$ in the document $d$. 
When a term appears more than once in a document, 
its first appearence is used. 
$k_{\tiny location,1}$ and $k_{\tiny location,2}$ are 
constants which are set according to the results of experiments. 

\item 
\underline{Categorical information ($K_{\tiny category}$)} 

$K_{\tiny category}$ uses category information 
such as whether or not the document appears 
on the economic or political pages. 
This operates by applying the technique called relevance feedback \cite{r-feedback}. 
Firstly, we specify the categories which occur 
in the top 15 documents of the first retrieval 
when $K_{\tiny category} = 1$. 
Then, we increase the scores of documents 
that are in majority or most-frequent categories. 
For example, 
the top 15 documents of the first retrieval were 
most often from the economic pages, 
we increase the scores of a documents from economic pages 
and decrease the scores of all documents 
from other sections of the newspaper. 
$K_{\tiny category}$ is expressed as follows; 

\begin{equation}
\footnotesize
  \label{eqn:men}
\displaystyle K_{\tiny category}(d) = 1 + k_{\tiny category} \frac{(Ratio A(d) - Ratio B(d))}{(Ratio A(d) + Ratio B(d))}
\end{equation}
where $Ratio A$ is the proportion of 
the top 100 documents in a given category 
on the first retrieval. 
$Ratio B$ is the proportion of that category in all the documents. 
The value of $K_{\tiny category}(d)$ is large 
when $Ratio A$ is large 
(the top 100 documents of the first retrieval 
frequently appear on the same pages as a document $d$.) 
and $Ratio B$ is small 
(few of the documents appear on the same pages as $d$). 
$k_{\tiny category}$ is a constant which 
is set according to the results of experiments. 
\end{enumerate}

\subsection{How terms are extracted}
\label{sec:extract_keyword}
Before being able to use Eq. (\ref{eqn:score}) in information retrieval, 
we must extract terms from a query. 
This section describes how this is done. 
With regard to term extraction, 
we considered the several methods listed below. 

\begin{enumerate}
\item 
\underline{Method of using only the shortest terms}

This is the simplest method. 
In the method, 
the query sentence is divided into short terms 
by using a morphological analyzer or a similar tool. 
All of the short terms are used in the retrieval process. 
The method used to divide the query sentence into short terms 
is described in Section \ref{sec:how_to_divide}. 

\item 
\underline{Method of using all term patterns}

In the first method the terms are too short. 
For example, ``enterprise'' and ``amalgamation'' 
would be used instead of ``enterprise amalgamation.''\footnote{Although 
this part of the paper deals only 
with retrieval from Chinese-language texts, 
and not English, 
we have used English examples 
for the benefit of this English-lanugae journal's readers. 
This method handles compound nouns 
and can be applied not only to Chinese 
but also to English. } 
We felt that ``enterprise amalgamation'' 
should be used along with the two short terms. 
Therefore, we decided to use both short and long terms. 
We call this the ``all term-patterns method.'' 
For example, when 
``enterprise amalgamation materialization'' was input, 
we used  
``enterprise'', ``amalgamation'', 
``materialization'', 
``enterprise amalgamation'', 
``amalgamation materialization'', 
and ``enterprise amalgamation materialization'' 
as terms for information retrieval. 
We felt that 
this method would be effective 
because it makes use of all term patterns. 
We also felt, however, that 
it is inequitable that only the three terms 
``enterprise,'' ``amalgamation,'' ``materialization,'' 
are derived from 
``... enterprise ... amalgamation ... materialization ...'',
while 
six terms are derived from 
``enterprise amalgamation materialization.'' 
We examined several methods of normalization in 
preliminary experiments, 
then decided to divide the weight of each term 
by $\sqrt{\frac{n(n+1)}{2}}$, 
where $n$ is the number of successive words. 
For example, in the case of ``enterprise amalgamation materialization'', 
$n = 3$. 

\item \underline{Method using a lattice}

Although the method of using all-term patterns effectively 
uses all patterns of terms, 
it needs to be normalized by using 
the adhoc equation $\sqrt{\frac{n(n+1)}{2}}$. 
We thus considered a method in which 
all term patterns are stored in a lattice. 
We used the patterns in the path 
with the highest score on Eq. (\ref{eqn:score}). 
(This method is almost the same as 
Ozawa's \cite{ozawa_nlp99_eng}. 
The differences are 
the fundamental equation for information retrieval, 
and whether or not a morphological analyzer is used.)

\begin{figure}[t]
  \begin{center}
  \begin{minipage}{7cm}
      \begin{center}
        \hspace*{-0.5cm}
        \includegraphics[height=4cm,width=7.5cm]{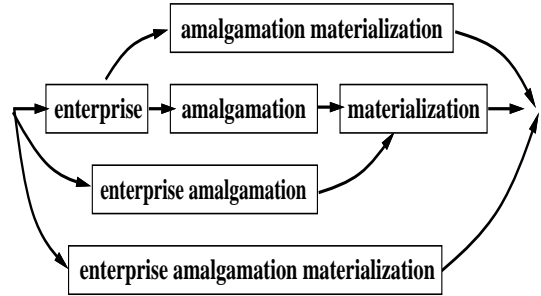} 
      \end{center}
      \vspace{-0.5cm}
    \caption{An example of a lattice structure}
    \label{fig:lattice}
    \end{minipage}
  \end{center}
\end{figure}

For example, 
in the case of ``enterprise amalgamation materialization'' 
the lattice shown in Fig. \ref{fig:lattice} is obtained. 
As shown in this figure, 
the score is calculated 
for each of the four paths 
by using Eq. (\ref{eqn:score}), 
and the terms in the highest-scoring path are used. 
This method does not require 
the adhoc normalization 
required by 
the method of using all term patterns. 

\item \underline{{Method of using} down-weighting {\cite{Fujita99_IREX}}}

This is the method that Fujita proposed at the IREX contest \cite{irex1_Sekine_eng}. 
It is similar to the all-term patterns method. 
It uses all term patterns but 
the method of normalization is different from 
that used in the all-term patterns method. 
The weights of the shortest terms are kept constant 
while the weights of the longer terms are decreased. 
We decided to apply the weight $k_{down}$$^{x-1}$ 
to such terms, 
where $x$ is the number of shortest terms and 
$k_{down}$ was set according to the results of experiments. 

\end{enumerate}

\subsection{The method dividing the query sentence into short terms}
\label{sec:how_to_divide}

We used the following three methods to 
divide the query sentence into short terms.\footnote{System A 
only segments sentences of documents 
are not segmented except for automatic feedback. } 
\begin{enumerate}
\item 
  \underline{Using a morphological analyzer}
  
  In this method, the query sentence is segmented by 
  using the CSeg\&Tag 1.0 
  Chinese-language morphological analyzer \cite{Sun_acl97}.

\item 
  \underline{Segmentation by using mutual information}

  This method is based on the method \cite{Sproat_acl97} proposed by 
  Sproat et al. 
  It calculates the mutual information of 
  two adjacent characters and divides them when 
  their mutual information. 
  The details of our method are as follows. 
  
  Almost all Chinese words consist of one Chinese character or 
  two Chinese characters.\footnote{According to 
    the paper \cite{Sproat_acl97}, 
    the occurrence rate of 
    words which consist of three Chinese characters is under 1\%.} 
  So we assumed that 
  all terms consist of one Chinese character or 
  two Chinese characters. 
  Thus, our method firstly divides Chinese sentences into fragments 
  which consist of one Chinese character or 
  two Chinese characters by using mutual information. 
  This is done by repeatly applying the following procedure. 
  \begin{itemize}
  \item 
    Divide up pairs of adjacent characters 
    with the lowest amount of mutual information, 
    where each pair is part of a fragment 
    which consist of more than two Chinese character. 
  \end{itemize}
  
  Next, we use the statistics of the Chinese corpus. 
  In this case, we assume that 
  the ratio of one-character words and two-characters words 
  in a Chinese text is a:b.\footnote{For example, Spraot stated that 
    this ratio is about 7:3 \cite{Sproat_acl97}.}
  We take this statistic then re-divide 
  those fragments that consist of pairs of characters 
  having little mutual information
  into two separate one-character words in such a way 
  that our process of division produces a text 
  broken up into one- and two-character words in the 
  approximate proportion a:b. 
  This is done by repeating the following procedure 
  until the text will be divided up to produce the 
  approximate proportion a:b. 
  \begin{itemize}
  \item 
    Divide those fragments 
    consisting of pairs of characters having the lowest mutual information
  \end{itemize}
  The result of this procedure is equivalent to 
  that of the following procedure. 
  \begin{itemize}
  \item 
    Divide up those fragments 
    consisting of pairs of characters having a level of mutual information 
    which is equal to or lower than $k_{cmi}$, 
    where $k_{cmi}$ is the amount of mutual information 
    that will divide up the text to produce the 
    approximate proportion a:b. 
  \end{itemize}

\item 
  \underline{Using both of the above two methods}

  This method firstly divides up the Chinese sentences by using 
  the morphological analyzer and then further divides up the fragments 
  by using mutual information and the statistics 
  on the Chinese corpus. 

\end{enumerate}

\subsection{Automatic feedback in System A}
\label{sec:feedback_in_system_b}

Automatic feedback is also used in System A. 
In System A, an element of automatic feedback is inclued via 
the IDF term of the equation (\ref{eqn:score}). 
When performing automatic feedback, 
we substitute the following equation 
for the original IDF term. 
{\footnotesize
\begin{eqnarray}
  \label{eqn:new_roccio}
IDF(t) & = &\{ E(t) + k_{af} \times (Ratio \ C(t) - Ratio \ D(t)) \} \nonumber \\
       &   & \times IDF_{orig}(t) 
\end{eqnarray}}
{\footnotesize
\begin{eqnarray}
  \label{eqn:new_roccio2}
E(t) & = & 1 \mbox{\ (when a term t is in a query)} \nonumber \\ 
  &   & 0 \mbox{\ (otherwise)}
\end{eqnarray}}
where $Ratio \ C(t)$ is the proportion of 
the top $k_{r}$ documents  of the first retrieval 
in which a term $t$ appears. 
$Ratio \ D(t)$ is the proportion of all of the documents 
in which a term $t$ appears. 
$IDF_{orig}(t)$ is the original IDF term. 
This formula is based on Rocchio's formula \cite{Roccio71}. 
$k_{af}$ and $k_{r}$ are constants set 
according to the results of experiments. 

Term expansion is also used in System A. 
The terms `Terms' as defined below are added. 
{\footnotesize
\begin{eqnarray}
  \label{eqn:new_test}
Terms = \{t| P(t) \geq k_{p}\}
\end{eqnarray}}
where $P(t)$ is the probability that 
a term $t$ appears in no less than $n$ documents 
of the top $k_{r}$ documents. 
$P(t)$ is approximately calculated by assuming that 
the appearance of the term $t$ follows 
a binominal distribution with a probability of 
the occurrence rate of the term $t$ in all the documents. 
$k_{p}$ is a constant set 
according to the results of experiments. 

\subsection{Weighting counting in automatic feedback}
\label{sec:weighting_counting}

We considered that 
a term which occurs in a document which has a higher rank 
on the first retrieval is more important. 
So, when counting the frequency of a term $t$ in a document d 
with a rank of $Rank(d)$, 
System A applied the following factor $AFW(t,d)$ to the frequency. 
{\footnotesize
\begin{eqnarray}
  \label{eqn:weighting_counting}
AFW(t,d) = (k_{afw}+1) - 2 \times k_{afw} \frac{Rank(d)-1}{k_{r}-1}
\end{eqnarray}}
where $k_{afw}$ is a constant set 
according to the results of experiments. 
Equations (\ref{eqn:new_roccio}) and (\ref{eqn:new_test}) are 
calculated by using the frequency calculated 
by Equation \ref{eqn:weighting_counting}. 

\begin{table*}[t]
\small
  \begin{center}
    \caption{Experimental results in CC Tasks}
    \begin{tabular}{|l|c|c|c@{ }c@{ }c@{ }c@{ }c@{ }c@{ }c@{ }c@{ }c|rr|rr|} \hline
  & &  & \multicolumn{9}{c|}{parameters} & \multicolumn{2}{c|}{R-Precsision} & \multicolumn{2}{c|}{Ave. Presision} \\\cline{4-16}
 & Task & ID  & Term & $k_{mi}$ & $k_{Nq}$ & dw & af & L & C & $k_{r}$ & $k_{af}$ & \multicolumn{1}{c|}{rigid} & \multicolumn{1}{c|}{relax} & \multicolumn{1}{c|}{rigid} & \multicolumn{1}{c|}{relax}\\\hline
 S1 & LO & 07 &  MI & 4.5 & 0 & y & y & y & y & 5 & 0.7 & 0.5751 & 0.6630 & 0.6348 & 0.7261\\ 
 S2 & LO & -- &  MI & 4.5 & 0 & y & n & y & y & 5 & 0.7 & 0.5529 & 0.6564 & 0.6186 & 0.7146\\ 
 S3 & LO & -- &  MI & 4.5 & 0 & n & n & y & y & 5 & 0.7 & 0.5660 & 0.6572 & 0.6183 & 0.7118\\ 
 S4 & LO & 08 &  MI & 4.5 & 1 & y & y & y & y & 5 & 0.7 & 0.5842 & 0.6692 & 0.6392 & 0.7362\\ 
 S5 & LO & 09 &  MI & 4.5 & t & y & y & y & y & 5 & 0.7 & 0.5803 & 0.6651 & 0.6386 & 0.7342\\ 
 S6 & LO & 02 &  MI &   3 & 0 & y & y & y & y & 5 & 0.7 & 0.5812 & 0.6685 & 0.6439 & 0.7326\\ 
 S7 & LO & 03 &  MI &   3 & 0 & y & n & y & y & 5 & 0.7 & 0.5632 & 0.6699 & 0.6329 & 0.7231\\ 
 S8 & LO & 04 &  MI &   3 & 0 & n & y & y & y & 5 & 0.7 & 0.5865 & 0.6684 & 0.6438 & 0.7325\\ 
 S9 & LO & 05 &  MI &   3 & 0 & n & n & y & y & 5 & 0.7 & 0.5587 & 0.6695 & 0.6329 & 0.7229\\ 
S10 & LO & 06 &  MI &   3 & 1 & y & y & y & y & 5 & 0.7 & 0.5782 & 0.6813 & 0.6459 & 0.7409\\ 
S11 & LO & 10 &  MI &   3 & t & y & y & y & y & 5 & 0.7 & 0.5780 & 0.6724 & 0.6427 & 0.7383\\ 
S12 & LO & 19 &  MI &   4 & 1 & y & y & y & y & 5 & 0.7 & 0.5814 & 0.6767 & 0.6407 & 0.7399\\ 
S13 & LO & -- &  MI &   4 & 1 & y & y & n & y & 5 & 0.7 & 0.5659 & 0.6704 & 0.6316 & 0.7334\\ 
S14 & LO & -- &  MI &   4 & 1 & y & y & y & n & 5 & 0.7 & 0.5916 & {\bf 0.6945} & {\bf 0.6567} & {\bf 0.7488}\\ 
S15 & LO & -- &  MI &   4 & 1 & y & y & n & n & 5 & 0.7 & 0.5778 & 0.6822 & 0.6530 & 0.7445\\ 
S16 & LO & 18 &  MI &   4 & 1 & y & y & y & y & 5 & 1   & 0.5900 & 0.6752 & 0.6415 & 0.7387\\ 
S17 & LO & 20 &  MI &   4 & 1 & y & y & y & y & 7 & 0.7 & 0.5746 & 0.6778 & 0.6388 & 0.7374\\ 
S18 & LO & 21 &  MI &   4 & 1 & y & y & y & y & 10& 0.7 & 0.5605 & 0.6741 & 0.6299 & 0.7316\\ 
S19 & LO & 11 &  MI &   4 & 1 & y & y & y & y & 15& 0.7 & 0.5743 & 0.6776 & 0.6265 & 0.7291\\ 
S20 & LO & 12 &  MI &   4 & 1 & y & y & y & y & 20& 0.7 & 0.5577 & 0.6767 & 0.6254 & 0.7268\\ 
S21 & LO & 13 &  MI &   4 & 1 & y & y & s & s & 5 & 0.7 & 0.5709 & 0.6703 & 0.6203 & 0.7271\\ 
S22 & LO & 14 & T+M &   4 & 1 & y & y & y & y & 5 & 0.7 & 0.5924 & 0.6810 & 0.6486 & 0.7413\\ 
S23 & LO & -- & TAG &   4 & 1 & y & y & y & y & 5 & 0.7 & {\bf 0.5936} & 0.6803 & 0.6501 & 0.7419\\ 
S24 & LO & 15 & T+M &   4 & 1 & y & n & y & y & 5 & 0.7 & 0.5820 & 0.6778 & 0.6388 & 0.7290\\ 
S25 & LO & 17 & T+M &   4 & 1 & n & y & y & y & 5 & 0.7 & 0.5712 & 0.6739 & 0.6341 & 0.7276\\ 
S26 & LO & 16 & T+M &   4 & 1 & n & n & y & y & 5 & 0.7 & 0.5557 & 0.6628 & 0.6165 & 0.7145\\ \hline 
S27 & SO & 02 &  MI &   4 & 1 & y & y & y & y & 5 & 0.7 & 0.5831 & {\bf 0.6817} & 0.6340 & 0.7368\\ 
S28 & SO & 03 & T+M &   4 & 1 & y & y & y & y & 5 & 0.7 & {\bf 0.5974} & 0.6766 & {\bf 0.6529} & {\bf 0.7376}\\ \hline 
S29 & VS & 02 &  MI &   4 & 1 & y & y & y & y & 5 & 0.7 & 0.5990 & {\bf 0.6788} & 0.6516 & 0.7387\\ 
S30 & VS & 03 & T+M &   4 & 1 & y & y & y & y & 5 & 0.7 & {\bf 0.6089} & 0.6749 & 0.6596 & 0.7397\\ 
S31 & VS & -- & T+M &   4 & 1 & y & y & n & y & 5 & 0.7 & 0.5893 & 0.6669 & 0.6468 & 0.7282\\ 
S32 & VS & -- & T+M &   4 & 1 & y & y & y & n & 5 & 0.7 & 0.6027 & 0.6781 & {\bf 0.6722} & {\bf 0.7454}\\ 
S33 & VS & -- & T+M &   4 & 1 & y & y & n & n & 5 & 0.7 & 0.5889 & 0.6636 & 0.6563 & 0.7350\\ 
S34 & VS & -- & TAG &   4 & 1 & y & y & y & y & 5 & 0.7 & 0.6086 & 0.6757 & 0.6604 & 0.7399\\ \hline 
S35 & TI & 02 &  MI &   4 & 1 & y & y & y & y & 5 & 0.7 & {\bf 0.4683} & {\bf 0.5923} & {\bf 0.4813} & {\bf 0.6239}\\ 
S36 & TI & 03 & T+M &   4 & 1 & y & y & y & y & 5 & 0.7 & 0.4651 & 0.5770 & 0.4793 & 0.6118\\\hline 
    \end{tabular}

The number of queries is 50. The number of documents is 132,173. 
    \label{tab:experiment_chir}
  \end{center}
\end{table*}

\subsection{Experiments}
\label{sec:exp_system_B}
The experimental results of System A are 
shown in Table \ref{tab:experiment_chir}.
``LO'', ``SO'', ``VS'', and ``TI'' indicate 
a long-query task, a short-query task, a very short query task, 
and a title-query task. 
The column ``ID'' indicates the system id in the NTCIR 2 contest. 
``--'' in ``ID'' indicates a system which was not submitted 
for the formal run of the NTCIR 2 contest. 
The column ``Term'' indicates 
the method used to divide the query sentence up into short terms. 
``TAG'', ``MI'', and ``T+M'' respectively indicate 
the use of the Chinese morphological analyzer, 
mutual information, and 
both the morphological analyzer and mutual information. 
$k_{cmi}$,\footnote{In the CHIR newspapers database, 
using $k_{cmi}$ = 5.33, 4.96, 4.56, 4.10, and 3.53
divides up the text to produce the 
approximate proportions of 7:3, 6.5:3.5, 6:4, 5.5:4.5, and 5:5. } 
$k_{Nq}$, $k_{r}$, and $k_{af}$ are set 
as in Table \ref{tab:experiment_chir}. 
``dw'', ``af'', ``L'' and ``C'' indicate 
the down-weighting method, automatic feedback method, 
locational information, and categorical information. 
``y'' in a column indicates the use of the method, and 
``n'' indicates that the method was not used. 
When we do not use the down-weighting method, 
we use the shortest-terms method 
as the method of extracting terms.\footnote{Our
previous work \cite{iral2000} had confirmed that 
the use of all term patterns is not a good method, 
and that 
even the simple method of using only the shortest terms 
can achieve good results.} 
The other parameters are set as follows: 
$k_{\tiny location,1} = 1.2$, 
$k_{\tiny location,2} = 0.1$,  
$k_{\tiny category} = 0.1$, 
$k_{t} = 1$, 
$k_{q} = \infty$, 
$k_{p} = 0.9$, and 
$k_{afw} = 0.5$. 
``s'' in ``L'' and ``C'' means the strong setting
where $k_{\tiny location,1} = 1.3$, 
$k_{\tiny location,2} = 0.15$,  
$k_{\tiny category} = 0.15$. 
``t'' in ``$k_{Nq}$'' means using $log\frac{Nq}{qf(t)}$ 
in a more complex way such that $qf(t)$ means 
the number of queries whose titles contain a term $t$. 

The following were the findings produced by the experimental results. 
\begin{itemize}
\item 
  The precisions of ``T+M'' or ``TAG'' are slightly higher than 
  that of ``MI.'' We thus found that 
  using the morphological analyzer produced better results 
  than using mutual information. 
  
\item 
  By comparing S12 with S13 or S30 with S31, 
  we found that locational information achieved 
  an improvement of about 0.02 or 0.03. 
  We can see that locational information is very effective. 
  
\item 
  By comparing S12 with S14 or S30 with S32, 
  we found that
  the precisions when categorical information not used were higher 
  than the precisions when it was used.  
  So, at least for these data, 
  using category information was not a good thing. 

\item 
  The automatic feedback method was always effective. 
  
\item 
  The down-weighting method sometimes produced better results and 
  sometimes produced poorer results. 

\end{itemize}

\subsection{Summary}

System A uses such characteristics of newspapers 
as locational information and obtained good results in the CC Tasks. 
By performing comparative experiments, 
we confirmed that locational information was effective. 
The other kinds of information were, however, not so effective. 

System A has many parameters and many methods. 
In the future, we would like to conduct much more extensive experiments 
in order to examine the effects of parameters and methods in System A. 


\section{Japanese and English IR Tasks}
\label{sec:je}

\subsection{Overview of the results}
\label{sec:ov}

The average precisions for System-B against relevant
documents on JJ, EJ, EE, and JE tasks are presented in Table
\ref{tab:ov}. In Table \ref{tab:ov}, `very short' means that
the system used the `TITLE' part of the queries for retrieval,
`short' means that it used the `DESCRIPTION' part of the queries,
and `long' means that it used all parts of the queries except
the `FIELD' part.  For each task, `feedback' means the
precisions that were obtained by automatic-feedback retrieval,
while `initial' means the precisions that were obtained by
using the raw initial queries. The symbol `$*$' means that the
corresponding search results from System-B were submitted to
the NTCIR 2 workshop committee as formal runs.\footnote{On JJ
  short, System-A outperformed System-B. Its best average
  precision was 0.3730} For the JJ and EE tasks, only
`feedback' results from System-B were submitted, while for the
EJ and JE tasks, both `initial' and `feedback' results were
submitted. These average precisions place the system in the
highest-scoring group among those for which results were submitted.

\begin{table*}[htbp]
  \begin{center}
    \caption{Average Precision (Relevant).}
    \begin{tabular}{|cc|ccc|}\hline
      &     & very short   & short        & long\\\hline
   JJ & \begin{tabular}{cc} initial \\ feedback \end{tabular}
      & \begin{tabular}{cc} 0.2112 \\ 0.2706$^*$ \end{tabular}
      & \begin{tabular}{cc} 0.3082 \\ 0.3396$^*$ \end{tabular}
      & \begin{tabular}{cc} 0.3807 \\ 0.4303$^*$ \end{tabular}\\\hline
   EJ & \begin{tabular}{cc} initial \\ feedback \end{tabular}
      & 
      & \begin{tabular}{cc} 0.2497$^*$ \\ 0.2564$^*$ \end{tabular}
      & \begin{tabular}{cc} 0.3156$^*$ \\ 0.3260$^*$ \end{tabular}\\\hline
   EE & \begin{tabular}{cc} initial \\ feedback \end{tabular}
      & \begin{tabular}{cc} 0.2192 \\ 0.2523$^*$ \end{tabular}
      & \begin{tabular}{cc} 0.2714 \\ 0.3131$^*$ \end{tabular}
      & \begin{tabular}{cc} 0.3684 \\ 0.4043$^*$ \end{tabular}\\\hline
   JE & \begin{tabular}{cc} initial \\ feedback \end{tabular}
      & 
      & \begin{tabular}{cc} 0.3409$^*$ \\ 0.3413$^*$ \end{tabular}
      & \begin{tabular}{cc} 0.3855$^*$ \\ 0.3856$^*$ \end{tabular}\\\hline
    \end{tabular}
    
    `*' represents submitted runs.
    
    \label{tab:ov}
  \end{center}
  
\end{table*}

We describe System-B in detail below. We start by describing
the scoring function used to rank documents. Next, we describe
the design issues involved in selecting possible free
parameters and then compare results for various parameter values through
experimented results. Finally, we conclude this section with a brief
summary.

\subsection{Scoring function}
\label{sec:sc}

Our scoring function is based on BM11
\cite{robertson92:_some_simpl_effec_approx_poiss}. Let $D$ be a
document and $Q$ be a query, where $D$ and $Q$ have been
tokenized into words. $D$ and $Q$ are bags of words. 
We define $|X|$ as the number of words in $X$ and define
$tf(w|X)$ as the number of a word $w$ in $X$. We also define
$W(X)$ as the set of different words in $X$.

The score of $D$ given $Q$, $\score(D|Q)$, is defined as:
\begin{equation}
  \label{eq:score}
  \score(D|Q) = \sum_{w \in W(D)\cap W(Q)} d(w|D) q(w|Q),
\end{equation}
where $d(w|D)$ is the weight of $w$ given $D$ and $q(w|Q)$ is
the weight of $w$ given $Q$. $d(w|D)$ is defined as:
\begin{equation}
  \label{eq:dwD}
  d(w|D) = \frac{\tf(w|D)}{\tf(w|D)+|D|/\Delta},
\end{equation}
where $\Delta$ is the average of $|D|$ over the document
collection $\D$ that contains $D$,i.e.,
\begin{equation}
  \Delta = \sum_{D\in\D} |D|/|\D|,
\end{equation}
where $|\D|$ is the number of documents in $\D$. $q(w|Q)$ is
defined as:
\begin{equation}
  \label{eq:qw}
  q(w|Q) = \frac{(k_q + 1)\tf(w|Q)}{k_q + \tf(w|Q)} \idf(w),
\end{equation}
where $k_q = 1000$ and 
\begin{equation}
  \idf(w) = \log \frac{|\D|}{|\D(w)|},
\end{equation}
where $|\D(w)|$ is the number of documents that contain $w$. 
$\D(w)$ is , of course, a subset of $\D$.

$\score(D|Q)$ is used for the initial search. For an automatic feedback
search, we use $\Score(D|Q)$:
\begin{equation}
  \label{eq:SDQ}
  \Score(D|Q) = \sum_{w \in W(D)\cap W(Q^\prime)} d(w|D) q^\prime(w|Q),
\end{equation}
where
\begin{equation}
  \label{eq:q2}
  q^\prime(w|Q) = \alpha q(w|Q) + \frac{\sum_{i=1}^{R} q(w|F(D_i))}{R},
\end{equation}
where $\alpha$ is a number , $D_i$ is the top $i$-th document retrieved
 by initial search, $R$ is the number of top-scoring documents
used in the automatic-feedback search, and $F$ is the function
used to select appropriate terms from a document. 
$Q^\prime$ in Equation (\ref{eq:SDQ}) is defined as:
\begin{equation}
  Q^\prime = Q \cup F(D_1) \cup \cdots \cup F(D_R).
\end{equation}

\subsection{Design Issues}
\label{sec:design}

The free parameters we consider in this paper are $\alpha$,
$F$, and $R$ in Equation (\ref{eq:q2}). We tried to have these
parameters  defined automatically. Before, however, we describe our attempts at
determining these parameters, we will discuss how we
preprocessed documents and queries for the JJ, EE, JE, and EJ
tasks.\footnote{The method used to preprocess documents and
  queries for CC tasks is similar to, but more primitive than,
  a method described in section \ref{sec:c}. We, thus,  omit a description
  here.}

\subsubsection{Tokenization}

Tokenization is, to a large degree, language dependent. 

We tokenized Japanese texts (documents or queries) by using
ChaSen version 2.02\footnote{\texttt{http://chasen.aist-nara.ac.jp/}}
\cite{matsumoto99:_japan_morph_analy_system_chasen_manual}
and then extracted lemmas of content words as $D$ or $Q$. We
postprocessed the output of ChaSen to eliminate some erroneous patterns of
tokenization.

In a similar way, we used LimaTK\footnote{\texttt{http://cl.aist-nara.ac.jp/\~{}tatuo-y/ma/}} to morphologically analyze English texts 
and then used a  stemmer that built around a library available in the WordNet1.6 package\footnote{\texttt{http://www.cogsci.princeton.edu/\~{}wn/}} to lemmatize content words.
Stop words were removed according to the list in the Nice stemmer package.\footnote{\texttt{http://www.ils.unc.edu/iris/irisnstem.htm}}

The documents and queries thus processed were used for the JJ and EE
tasks.

\subsubsection{Query translation}

For the JE and EJ tasks, we translated queries. Once we translate
queries,  cross-lingual IR (CLIR, i.e., JE or EJ) is
performed by the same method as used for mono-lingual IR (JJ or
EE). We describe the method below as applied to the translation of 
a Japanese query into English. English to Japanese translation
is performed in a similar way.

We perform document expansion
\cite{singhal99:_docum_expan_speec_retriev} to augment the original
queries; i.e., for a Japanese query, we first search the
Japanese database to get documents that are relevant to the query. Next,
we extract the words contained in the top-5 documents and
combine them to the original query. We thus obtain an expanded
Japanese query.\footnote{Local context analysis has been used
  to expand queries in CLIR
  \cite{ballesteros97:_phras_trans_query_expan_techn}. The
  comparison is a future work.}

The expanded Japanese query is then translated into English. 
For the translation, we first made a Japanese-to-English
bilingual dictionary from the Japanese-English abstract pairs
provided for the first NTCIR Workshop. From those pairs, we
extracted Japanese-English keyword pairs contained in the
abstract pairs. It was possible for these keywords to be phrases or
words. If a Japanese keyword co-occurred with multiple English
keywords, then we selected the most frequently co-occurring English
keyword as the translation of the Japanese keyword
\cite{chen99:_compar_japan_japan_englis}. 
Texts were translated in the following two steps;  we used ChaSen to
morphologically analyze the text, then 
translated the sequence of morphemes  into English. The translation
was  on a word-to-word or phrase-to-phrase basis. 
Disambiguation by contexts was not used. The translation was based on
longest matches. For example, if a query `a b c' is given,
where `a' is translated into `A' and `a b c' is
translated into `D E', then `a b c' is translated into
`D E'.\footnote{\cite{chen99:_compar_japan_japan_englis} also
  used a longest-match algorithm, but they did not use a
  morphological analyzer, which might degrade the system
  performance. This belief is supported by Table \ref{tab:clir}
  which shows the performance of our method in no document
  expansion. The average precision of
  \cite{chen99:_compar_japan_japan_englis} on the same task
  was 0.3216, while that of our approach is 0.3364.}

Translated queries were used for the JE and EJ tasks. The retrieval
algorithm was the same as that used for the JJ and EE tasks.

As is shown in Table \ref{tab:ov}, our approach to the JE and EJ tasks
worked quite well. It is evident, however, that the degree of
success of our approach depends on the degree of  similarity
between the Japanese database and the English database used for
CLIR. We thus conducted another experiment which used the
databases and JE-queries provided for the first NTCIR Workshop. The
type of query used for the experiment was `long' except that we
did not use English concepts.

\begin{table}[htbp]
  \begin{center}
    \caption{Average precisions with document expansion.}
    \begin{tabular}{|ccc|} \hline
      Source & Target & Average precision \\  \hline
          $\phi$        & ntc1-e           & 0.3364\\
      ntc2-j            & ntc1-e           & 0.3628\\
      ntc1-j            & ntc1-e           & 0.3899\\ \hline
    \end{tabular}
    \label{tab:clir}
  \end{center}
\end{table}

In Table \ref{tab:clir}, the column `Source' lists the
databases used to expand the original queries. `$\phi$'
indicates no document expansion. `ntc2-j' means that the
Japanese database which was freshly added for the second NTCIR
Workshop was used for document expansion, and `ntc1-j' means
that the Japanese database provided for the NTCIR workshop 1
was used for document expansion. `ntc1-e', which is listed in
`Target' column for all entries, is the English database that
was the target of the searches for documents. Average precision
was evaluated against relevant documents in `ntc1-e'.

`ntc1-j' and `ntc1-e' are nearly parallel. Naturally, it
achieved the best performance of these three cases. 
`ntc2-j' and `ntc1-e' are comparable. The average precision is
still better than with no document expansion. Document
expansion is thus worthwhile for CLIR.

We have briefly described the language-dependent parts of System-B.
Next, we describe its language-independent parts, describing $F$,
$R$, and $\alpha$ in Equation (\ref{eq:q2}), in that order.

\subsubsection{Definition of $F$}

We define a relevance of word $w$ for the top-scoring $R$ documents in terms
of probability.\footnote{\cite{robertson99:_okapi_keenb_trec}
  also uses a probabilistic metric to select relevant terms.}

Given a bag of words $X$, then the probability of $w$,
$\Pr(w|X)$, and its variance $\Var(w|X)$ are estimated as
\begin{equation}
  \Pr(w|X) = \frac{\tf(w|X)+1}{|X|+2},
\end{equation}
\begin{equation}
  \Var(w|X) = \frac{\Pr(w|X)*(1-\Pr(w|X))}{|X|+3}.
\end{equation}
We then define $D^1_R$ as the bag of words that contains all
the words in $D_1, D_2, \cdots, D_R$ and define
$\overline{D^1_R}$ as the complement of $D^1_R$ with a universal
set that is defined by all of the words in the document collection
$\D$.

The relevance of word $w$, $\rel(w|D^1_R)$, is defined as
\begin{equation}
  \rel(w|D^1_R) = \frac{\Pr(w|D^1_R)-\Pr(w|\overline{D^1_R})}{\sqrt{\Var(w|D^1_R)+\Var(w|\overline{D^1_R})}}.
\end{equation}
Finally we define $F(D_i)$ as
\begin{equation}
  F(D_i) = \{w | \rel(w|D^1_R) >= \theta \wedge w \in D_i\},
\end{equation}
where $\theta$ is a predefined threshold.

$\rel(w|D^1_R)$ approximately follows the standard normal
distribution. Possible candidates for $\theta$ are 1.28, 1.65
and 2.33, which correspond to significance levels of 0.10,
0.05 and 0.01, respectively. Hereafter, significance levels are
represented by $p$.

We used $p=0.10 (\theta=1.28)$ for all of the submitted
runs.\footnote{We used a more precise value for $\theta$
  actually.} This choice was based on previous experiments conducted on the
database provided for the first NTCIR Workshop.  $p=0.10$ is a
robust parameter value for term selection as is shown in section
\ref{sec:comp}.

\subsubsection{Definition of $R$}

We used the method explained below to set $R$ automatically. We
found, however, that the method was not efficient, and this is shown in section
\ref{sec:comp}.

Our method is based on the degree of increase in the number of
different words in top-scoring documents. If the content of
successive documents is similar, the documents should
share keywords. This degree of increase is thus low when
similar documents continue.  Our algorithm is depicted in
Figure \ref{fig:al}. In the experiments described in section
\ref{sec:comp}, the average numbers of documents selected by the
algorithm were 3.81, 4.01, and 3.95, for `very short', `short',
and `long' queries, respectively.

As is shown in section \ref{sec:comp}, the performance of IR is
quite sensitive to $R$. We will therefore investigate methods
for the automatic-determination of $R$ in future work, though our
initial attempts have not been too successful.

\begin{figure}[htbp]
  \begin{center}
    \begin{tabular}{|l|}
\hline
\begin{minipage}{20em}

\medskip
\tt
\begin{verbatim}
for(R=3;;R++){
  if (diff(R) > diff(R-1)) {
    break
  }
}
\end{verbatim}

\noindent
int diff(i) \{

\hspace*{2em}return $|W(F(D^1_i))| - |W(F(D^1_{i-1}))|$

\}

\medskip

\end{minipage}
\\ \hline
\end{tabular}
\caption{Algorithm for determining $R$.}
    \label{fig:al}
  \end{center}
\end{figure}

\subsubsection{Definition of $\alpha$}

$\alpha$ is defined heuristically as follows:
\begin{equation}
  \label{alpha}
  \alpha = |W(F(D^1_R))|^{\frac{1}{|W(Q)|}},
\end{equation}
where 
\begin{equation}
  F(D^1_R) = F(D_1) \cup F(D_2) \cup \cdots \cup F(D_R).
\end{equation}
$\alpha \ge 1 $ holds because $|W(F(D^1_R))| \ge 1$. $\alpha$
approaches 1 when $|W(Q)|$ is large. $\alpha$ takes a large
value when the number of different words in $Q$ is small and
the number of different words in $D^1_R$ is large. $\alpha$
is defined so that $Q$ is more important than $D^1_R$. In the experiments
described in section \ref{sec:comp}, the average value of $\alpha$
were 13.44, 3.89, and 1.14, for `very short', `short', and
`long' queries, respectively.

This heuristic approach worked reasonably well as is shown in section
\ref{sec:comp}.

In summary, for the formal runs, we used $p=0.10$ for term
selection and used automatic methods to set $R$ and $\alpha$. 
$p=0.10$ was the only parameter that we had to set by hand.

\subsection{Comparison of parameter values}
\label{sec:comp}

We varied the values of $p$, $R$, and $\alpha$ to observe the
effects of parameter values on performance. Performance was
measured by the average precision against relevant documents. 
We used the queries and documents provided for the second NTCIR
workshop. Experiments were conducted on JJ and EE tasks.  
We  only report on the results for JJ tasks, here, because both sets of results
displayed the same tendency.

The parameter values for $p$ were
\begin{equation}
  p = 0.10, 0.05, 0.01.
\end{equation}
The parameter values for $R$ were
\begin{equation}
  R = 1, 3, 5, 7, 10, 15.
\end{equation}
The parameter values for $\alpha$ were
\begin{equation}
  \alpha = 0.5, 1.0, 1.5.
\end{equation}
For $R$ and $\alpha$, we also tried the heuristic methods
described in Figure \ref{fig:al} and Equation (\ref{alpha}). We
tried all combinations of these parameter values. Thus, we
conducted $3 \times 7 \times 4 = 84$ runs to make our
comparison for each of the `long', `short', and `very short'
queries.

\begin{figure*}[htbp]
  \begin{center}
    \includegraphics[scale=.7]{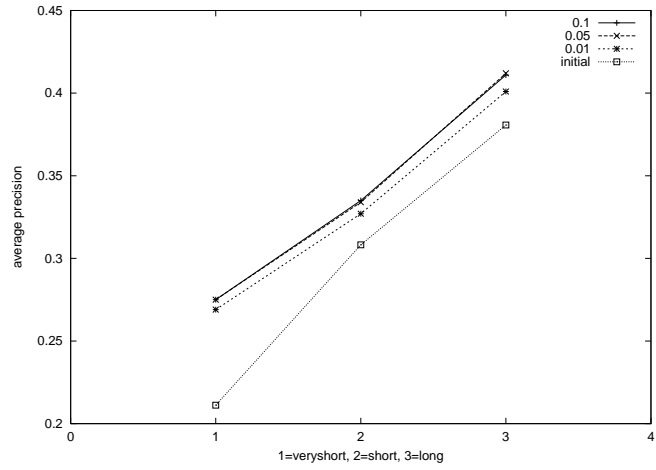}
    \caption{Average precisions for various settings of $p$}
    \label{fig:p}
  \end{center}
\end{figure*}

\begin{figure*}[htbp]
  \begin{center}
    \includegraphics[scale=.7]{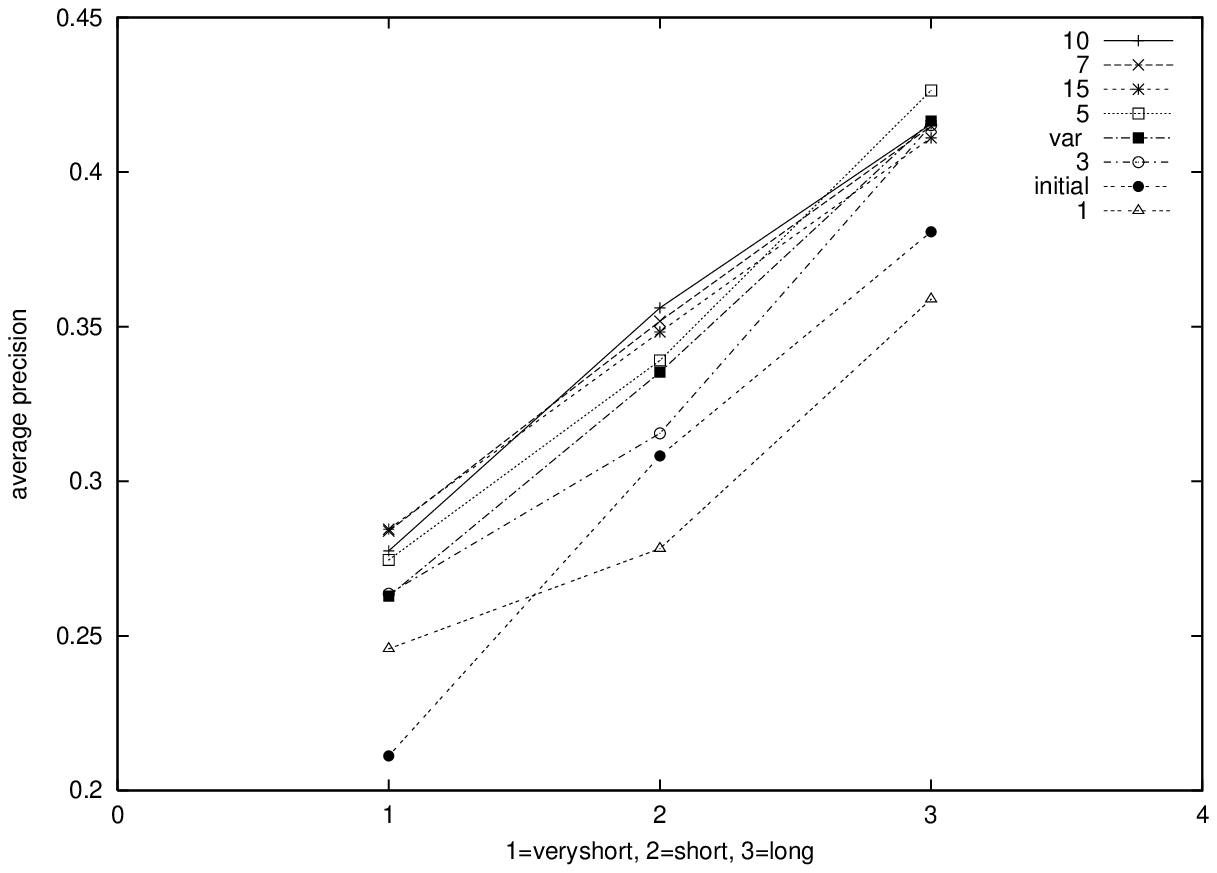}
    \caption{Average precisions for various settings of $R$}
    \label{fig:R}
  \end{center}
\end{figure*}

\begin{figure*}[htbp]
  \begin{center}
    \includegraphics[scale=.7]{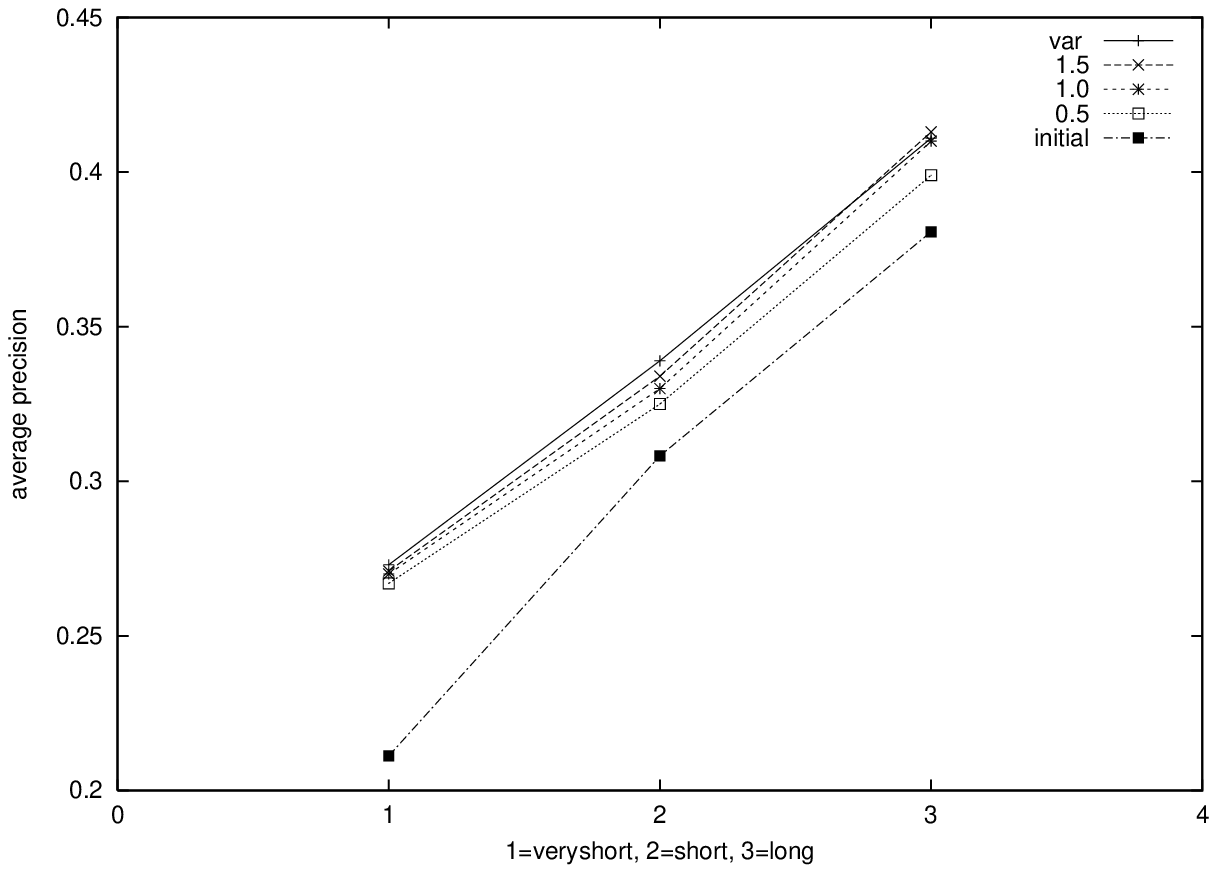}
    \caption{Average precisions for various settings of $\alpha$}
    \label{fig:alpha}
  \end{center}
\end{figure*}

To evaluate the effectiveness of a parameter, we fixed its
value and then calculated the average of the average precisions
of the 84 runs. The results are shown in Figures \ref{fig:p},
\ref{fig:R}, and \ref{fig:alpha}. In these figures, horizontal
axes represent the query types and vertical axes represent the
average precisions. The title of each line indicates the
parameter value. `var' means that values are determinbe by our
methods proposed above. `initial' means the results for the initial
search. The titles are in order of decreasing
 average precision for short queries.

Figure \ref{fig:p} shows the results for various settings of
$p$. Note that $p=0.1$ and $p=0.05$ performed equally well. 
This suggests that the value of $p$ is robust over this
range.\footnote{Additional experiments showed that average
  precisions for $p=0.9$ to $0.05$ performs equally well. (The
  results of $p=0.1$ were slightly better than those of other
  values.)}

Figure \ref{fig:R} shows the results for various settings of
$R$. It is difficult to detect any clear tendency in Figure
\ref{fig:R}, but it seems that when queries are long, small $R$ values
perform well, and when queries are short, large $R$ performs
well. This suggests that the length of queries
could be used to set $R$ automatically.

Figure \ref{fig:alpha} shows the results for various settings
of $\alpha$. The average of $\alpha$ were 13.44, 3.89, and 1.14
for `very short', `short', and `long' queries, respectively.
$\alpha$ takes large values for `very short' and `short'
queries. It takes small values for `long' queries. $\alpha$
worked reasonably well.  This is because for `very short' and
`short' queries, the results of the initial search are not very
reliable, so we had to weight $Q$ heavily, while for `long'
queries, the results of the initial search are reliable, so we
don't have to weight $Q$ so heavily.

\subsection{Summary}
\label{sec:sum}

System-B was designed as a portable IR system that avoids free
parameters as much as possible. It will be possible to improve
the sysmtem's performance by providing a proper
method for determining the number of top-ranked documents to be used
in automatic-feedback.

\section{Conclusion}
\label{sec:conclusion}

We have developed two systems for the second NTCIR Workshop IR
tasks. One was 
an improved version of the system that
was used for the first NTCIR Workshop IR tasks and the IREX
Workshop IR tasks.
The other was 
a freshly developed system
that avoids free parameters as much as possible. 
The former system participated in the JJ and CC tasks and
the latter system participated in the JJ, EE, JE, EJ and CC tasks.
Both systems achieved good results.
We have not yet compared the two systems thoroughly.
In the future, we will conduct a more detailed examination of
our systems and will determine what kinds of information are effective. 

\bibliographystyle{latex8}
\bibliography{mysubmit}

\end{document}